\documentclass[conference]{IEEEtran}

\usepackage[utf8]{inputenc}

\usepackage{newtxtext,newtxmath}

\usepackage{graphicx}
\usepackage{amsmath}
\usepackage{booktabs}
\usepackage[table]{xcolor}
\usepackage{multirow}
\usepackage{tabularx}
\usepackage{placeins}
\usepackage{csquotes}
\usepackage{textcase}
\usepackage{float}
\usepackage{array}
\usepackage{pifont}
\usepackage{cite}
\usepackage{balance}
\usepackage[hidelinks]{hyperref}  

\newcolumntype{C}[1]{>{\centering\arraybackslash}m{#1}}
\newcolumntype{Y}{>{\centering\arraybackslash}X} 
\newcolumntype{L}{>{\raggedright\arraybackslash}p{2cm}} 
\newcolumntype{M}{>{\centering\arraybackslash}p{1.9cm}}  
\newcolumntype{Z}{>{\centering\arraybackslash}p{1.2cm}}  
\newcommand{\cmark}{\ding{51}} 
\newcommand{\xmark}{\ding{55}} 

\makeatletter
\renewenvironment{abstract}{%
  \par\vspace{0pt}\noindent\hspace*{2em}%
  {\bfseries\itshape Abstract}\textemdash\ \bfseries\ignorespaces
}{%
  \par\vspace{0.28\baselineskip}%
}
\renewenvironment{IEEEkeywords}{%
  \par\vspace{0pt}\noindent\hspace*{2em}%
  {\bfseries\itshape Keywords}\textemdash\ \bfseries\ignorespaces
}{%
  \par
}
\newenvironment{keywords}{\begin{IEEEkeywords}}{\end{IEEEkeywords}}
\makeatother

\providecommand{\ninept}{}

\title{\NoCaseChange{Beyond Pixels: A Training-Free, Text-to-Text Framework for Remote Sensing Image Retrieval}}
\author{
\IEEEauthorblockN{Jinghao Xiao}
\IEEEauthorblockA{School of Computer Science\\ Faculty of Engineering and Information Technology \\
University of Technology Sydney\\
Sydney, Australia\\
Jinghao.Xiao@student.uts.edu.au}
\and
\IEEEauthorblockN{Yiheng Guo}
\IEEEauthorblockA{School of Computer Science\\ Faculty of Engineering and Information Technology \\
University of Technology Sydney\\
Sydney, Australia\\
Yiheng.Guo@student.uts.edu.au}
\and
\IEEEauthorblockN{Xing Zi}
\IEEEauthorblockA{School of Computer Science\\ Faculty of Engineering and Information Technology \\
University of Technology Sydney\\
Sydney, Australia\\
Xing.Zi-1@uts.edu.au}
\and
\IEEEauthorblockN{Karthick Thiyagarajan}
\IEEEauthorblockA{Smart Sensing and Robotics Laboratory (SensR Lab)\\ Centre for Advanced Manufacturing Technology\\
Western Sydney University\\
Sydney, Australia\\
K.Thiyagarajan@westernsydney.edu.au}
\and
\IEEEauthorblockN{Catarina Moreira}
\IEEEauthorblockA{The Data Science Institute\\ Faculty of Engineering and Information Technology \\
University of Technology Sydney\\
Sydney, Australia\\
Catarina.PintoMoreira@uts.edu.au}
\and
\IEEEauthorblockN{Mukesh Prasad}
\IEEEauthorblockA{School of Computer Science\\ Faculty of Engineering and Information Technology \\
University of Technology Sydney\\
Sydney, Australia\\
Mukesh.Prasad@uts.edu.au}
}

\begin{document}
\ninept
\maketitle

\begin{abstract}
\textbf{Semantic retrieval of remote sensing (RS) images is a critical task fundamentally challenged by the \textquote{semantic gap}, the discrepancy between a model's low-level visual features and high-level human concepts. While large Vision-Language Models (VLMs) offer a promising path to bridge this gap, existing methods often rely on costly, domain-specific training, and there is a lack of benchmarks to evaluate the practical utility of VLM-generated text in a zero-shot retrieval context. To address this research gap, we introduce the Remote Sensing Rich Text (RSRT) dataset, a new benchmark featuring multiple structured captions per image. Based on this dataset, we propose a fully training-free, text-only retrieval reference called TRSLLaVA. Our methodology reformulates cross-modal retrieval as a text-to-text (T2T) matching problem, leveraging rich text descriptions as queries against a database of VLM-generated captions within a unified textual embedding space. This approach completely bypasses model training or fine-tuning. Experiments on the RSITMD and RSICD benchmarks show our training-free method is highly competitive with state-of-the-art supervised models. For instance, on RSITMD, our method achieves a mean Recall of 42.62\%, nearly doubling the 23.86\% of the standard zero-shot CLIP baseline and surpassing several top supervised models. This validates that high-quality semantic representation through structured text provides a powerful and cost-effective paradigm for remote sensing image retrieval.
}
\end{abstract}


\begin{figure*}[t!]  
  \centering
  \makebox[\textwidth][c]{%
    \includegraphics[width=0.88\textwidth]{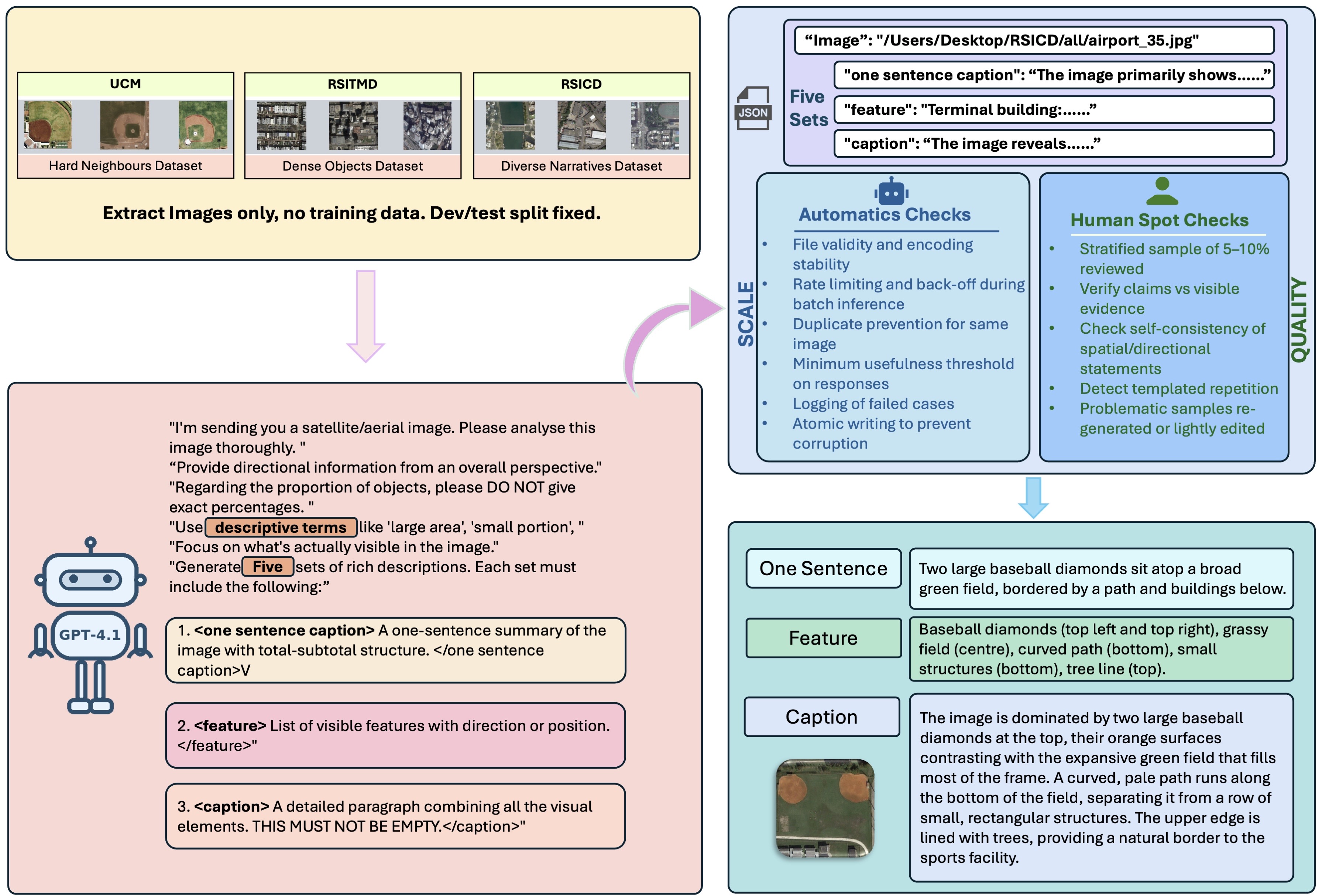}    %
  }
  \caption{Overview of the RSRT dataset construction pipeline.}
  \label{fig:rsrt-pipeline}
  \label{fig:flowchart} 
\end{figure*}

\section{Introduction}
With the rapid advancements in satellite technologies and sensor capabilities, Earth observation has entered an era of unprecedented data explosion~\cite{li2021image, li2024vision, al2022multilanguage}. Remote sensing image archives are expanding at an extraordinary pace, presenting tremendous opportunities for applications such as military reconnaissance, environmental monitoring, and urban planning~\cite{li2021image, li2024vision}. However, this explosive growth also poses significant challenges in data management, particularly in information discovery. The ever-growing repositories create an urgent need for efficient and accurate retrieval of relevant content. To address this, Remote Sensing Image Retrieval (RSIR) has emerged, with its core task being to search and return semantically relevant images from large-scale databases~\cite{barz2021content, li2021image}.

Initially, RSIR systems relied on Content-Based Image Retrieval (CBIR), which matched images based on low-level visual features like color, texture, and shape~\cite{li2021image, yuan2021lightweight, yuan2022remote}. These methods were fundamentally limited by the \textquote{semantic gap} which is the discrepancy between pixel level appearance and high level human concepts~\cite{li2021image}. To bridge this gap, modern approaches have shifted towards Vision-Language Models (VLMs) like CLIP, which learn to map images and text into a shared semantic space, enabling powerful cross modal retrieval~\cite{radford2021learning, li2024vision}. Despite their success in general domains, directly applying these VLMs to remote sensing reveals a distinct set of challenges rooted in the data's unique nature~\cite{li2024vision, li2023rs, liu2024remoteclip}.

Meanwhile, the community has developed image–text benchmarks in RS to facilitate cross-modal research, such as RSICD~\cite{lu2017exploring}, RSITMD~\cite{yuan2022exploring}, and UCM~\cite{yang2010bag}, which are widely used for remote sensing captioning and retrieval. Recent RS-tailored VLM efforts (e.g., RemoteCLIP) further adapt contrastive language–image pretraining to satellite imagery and inspire training paradigms for zero-shot or low-shot transfer~\cite{liu2024remoteclip, li2023rs, zhang2024rs5m}. However, this dominant paradigm of fine-tuning large models on domain-specific datasets introduces its own fundamental limitations that hinder true progress. This paper identifies three key challenges in existing methods: (1) The lack of true zero-shot generalization. Many so called zero-shot models have undergone extensive supervised training. They perform well on their specific datasets but fail when they encounter new scenes or different types of images. For example, a model trained on summer images might not recognize the same location in winter. They lack the ability to work consistently in a changing environment. (2) Coarse-grained image representation. Existing methods typically compress a complex remote sensing image into a single global feature vector. This process loses all the small but critical details. As a result, users cannot perform fine-grained queries, such as searching for \textquote{a sports field with a red running track} or \textquote{blue cars in a parking lot}. The system can only understand the general overview of an image, not its specific contents. (3) An implicit and entangled feature space. The single feature vector not only loses details but also mixes all semantic information together, such as objects, attributes, and spatial relationships. This form of representation is implicit and cannot be separated. This makes it impossible for users to query specific attributes or relationships, limiting retrieval to coarse, scene level matching.

To address this, we introduce a training-free framework that sidesteps these issues by reformulating cross-modal retrieval as a pure text-to-text (T2T) matching problem. Our key insight is that instead of trying to align noisy image features with text, we can leverage a SOTA VLM to convert the entire image database into a corpus of rich, structured textual descriptions. This offline process creates a high quality, searchable semantic database. All subsequent retrieval tasks, whether starting from a text or image query, are then performed entirely within a unified textual embedding space. This approach not only eliminates the need for any training or fine-tuning but also naturally supports fine-grained, compositional queries through the explicit nature of text.

The main contributions of this work are summarized as follows:
\begin{itemize}
    \item \textbf{A Novel Training-Free Paradigm:} We introduce TRSLLaVA, a new framework for RSIR that completely avoids domain-specific training. By converting all images into rich text, we reformulate retrieval as a more robust and efficient text-to-text matching problem, offering a true zero-shot solution.

    \item \textbf{The RSRT Dataset for Fine-Grained Evaluation:} We construct and release the Remote Sensing Rich Text (RSRT) dataset, the first benchmark specifically designed to evaluate fine-grained, rich-text-based retrieval, providing multiple structured caption variants for each image.

    \item \textbf{State-of-the-Art Zero-Shot Performance:} Through extensive experiments, we demonstrate that our training-free method is not only highly competitive with fully supervised models but also nearly doubles the performance of the standard zero-shot CLIP baseline, validating our approach as a powerful and cost-effective alternative to the dominant fine-tuning paradigm.
\end{itemize}

\section{Methodology}

Our methodology introduces a paradigm shift for remote sensing image retrieval. Instead of relying on complex and often noisy cross-modal alignment between images and text, we reformulate all retrieval tasks into a simpler, more robust text-to-text matching problem. This approach is motivated by two key insights. First, modern Vision-Language Models (VLMs), derived from Large Language Models (LLMs), excel at generating long, descriptive, and information-rich text, far surpassing the short, label-like sentences produced by older NLP models~\cite{liu2023visual,dong2024benchmarking}. Second, retrieval frameworks analogous to Retrieval-Augmented Generation (RAG) perform significantly better when operating on these detailed, rich-text documents rather than short, ambiguous phrases~\cite{gao2022precise}.

Following this principle, our framework first transforms all images into high-quality textual representations. The entire process begins with the construction of our rich-text corpus, as illustrated in Figure~\ref{fig:rsrt-pipeline}. The process starts with images sourced from established remote sensing benchmarks, including RSITMD and RSICD. Each image is then processed by an  LLM model (in this case, GPT-4.1) using a carefully designed, instruction-driven prompt. This prompt guides the model to generate five distinct sets of rich textual descriptions for every image, enforcing a structured output that includes a concise one-sentence summary, a list of directional and relational features, and a detailed descriptive paragraph.

The raw outputs undergo a rigorous two-stage quality control process to ensure both scale and quality. First, a series of automated checks are performed to handle issues like file validity, de-duplication, and response formatting. Following this, a human auditing phase involves spot-checking a sample of the descriptions to verify factual claims and check for internal consistency. The final, validated output is a high-quality, structured set of textual annotations for each image, which collectively form the RSRT dataset. This rich-text corpus serves as the foundation for our subsequent training-free retrieval framework, where all retrieval tasks, whether starting from a text or image query, are performed entirely within a unified textual embedding space.

\subsection{RSRT: Building a Fine-Grained Rich-Text Corpus}

The foundation of our framework is a high-quality semantic database. Let the source dataset be a collection of $N$ images, $\mathcal{I} = \{I_1, I_2, \dots, I_N\}$. To construct our corpus, we introduce the Remote Sensing Rich Text (RSRT) dataset. We process standard benchmarks (RSICD, RSITMD) using a state-of-the-art LLM, GPT-4.1\cite{openai_gpt4tr_pdf}, which is chosen for its unparalleled ability to generate detailed and structured descriptions.

For each image $I_i \in \mathcal{I}$, we use carefully designed prompts, $P_{\text{structured}}$, to elicit a set of five structured caption variants from GPT-4.1. This set, denoted as $\mathcal{D}_i$, contains rich textual representations of the image:
\begin{equation}
\mathcal{D}_i = \{ D_{i,v} \}_{v=1}^5 = f_{\text{GPT-4.1}}(I_i, P_{\text{structured}})
\label{eq:rsrt_generation}
\end{equation}
where $D_{i,v}$ is the $v$-th caption variant for image $I_i$. These variants include: (i) a concise one-sentence summary; (ii) a directional and relational feature list; and (iii) a detailed descriptive paragraph. The final RSRT corpus, $\mathcal{C}_{\text{RSRT}}$, is the collection of all image-description pairs:
\begin{equation}
\mathcal{C}_{\text{RSRT}} = \{ (I_i, \mathcal{D}_i) \}_{i=1}^N
\label{eq:rsrt_corpus}
\end{equation}
This multi-variant structure ensures that nuanced, fine-grained information is explicitly preserved, serving as the rich semantic database for retrieval.

\begin{table*}[t]
\centering
\scriptsize
\setlength{\tabcolsep}{3pt}
\caption{RESULTS FOR IMAGE-TO-TEXT AND TEXT-TO-IMAGE RETRIEVAL TASKS ON RSITMD.}
\label{tab:rsitmd_results}
\begin{tabularx}{\textwidth}{L M Z *{6}{Y} c}
\toprule
\multirow{2}{*}{\textbf{Method}} & \multirow{2}{*}{\textbf{Training Dataset}} & \multirow{2}{*}{\textbf{Zero-Shot}} & \multicolumn{3}{c}{\textbf{image\_to\_text}} & \multicolumn{3}{c}{\textbf{text\_to\_image}} & \multirow{2}{*}{\textbf{mR}} \\
\cmidrule(l{.7em}r{.7em}){4-6} \cmidrule(l{.7em}r{.7em}){7-9}
& & & recall@1 & recall@5 & recall@10 & recall@1 & recall@5 & recall@10 & \\
\midrule
LW-MCR~\cite{yuan2021lightweight} & RSITMD & \xmark & 10.18 & 28.98 & 39.82 & 7.79 & 30.18 & 49.78 & 27.79 \\
VSE++~\cite{faghri2017vse++} & RSITMD & \xmark & 10.28 & 27.65 & 39.60 & 7.96 & 24.87 & 36.87 & 24.54 \\
AMFMN~\cite{yuan2022exploring} & RSITMD & \xmark & 11.06 & 29.02 & 38.72 & 9.96 & 34.03 & 52.96 & 29.29 \\
SWAN & RSITMD & \xmark & 13.35 & 32.55 & 45.11 & 11.24 & 40.49 & 60.60 & 33.89 \\
GaLR~\cite{yuan2022remote} & RSITMD & \xmark & 14.82 & 31.64 & 42.48 & 11.15 & 36.68 & 51.68 & 31.41 \\
HVSA~\cite{zhang2023hypersphere} & RSITMD & \xmark & 13.20 & 32.08 & 45.58 & 11.43 & 39.20 & 57.45 & 33.15 \\
FAAMI~\cite{zheng2023fine} & RSITMD & \xmark & 19.32 & 35.62 & 48.89 & 12.96 & 42.39 & 59.95 & 36.52 \\
PIR~\cite{pan2023prior} & RSITMD & \xmark & 17.64 & 41.15 & 53.82 & 12.77 & 41.68 & 63.41 & 38.41 \\
Multilanguage~\cite{al2022multilanguage} & RSITMD & \xmark & \textbf{19.69} & 40.26 & 54.42 & 17.61 & \textbf{49.73} & 66.59 & 41.38 \\
MTGFE~\cite{zhang2023fusion} & RSITMD & \xmark & 17.92 & 40.93 & 53.32 & 16.59 & 48.50 & 67.43 & 40.78 \\
GeoRSCLIP~\cite{zhang2024rs5m} & RS5M & \xmark & 19.03 & 34.51 & 46.46 & 14.16 & 42.39 & 57.52 & 35.68 \\
CLIP-Baseline~\cite{radford2021learning} & - & \cmark & 9.53 & 21.03 & 32.74 & 8.81 & 27.85 & 43.19 & 23.86 \\
\textbf{OURS} & - & \cmark & 19.54 & \textbf{44.91} & \textbf{61.28} & \textbf{18.26} & 43.32 & \textbf{68.41} & \textbf{42.62} \\
\bottomrule
\end{tabularx}
\end{table*}

\begin{table*}[t]
\centering
\scriptsize
\setlength{\tabcolsep}{3pt}
\caption{RESULTS FOR IMAGE-TO-TEXT AND TEXT-TO-IMAGE RETRIEVAL TASKS ON RSICD.}
\label{tab:rsicd_results}
\begin{tabularx}{\textwidth}{L M Z *{6}{Y} c}
\toprule
\multirow{2}{*}{\textbf{Method}} & \multirow{2}{*}{\textbf{Training Dataset}} & \multirow{2}{*}{\textbf{Zero-Shot}} & \multicolumn{3}{c}{\textbf{image\_to\_text}} & \multicolumn{3}{c}{\textbf{text\_to\_image}} & \multirow{2}{*}{\textbf{mR}} \\
\cmidrule(l{.7em}r{.7em}){4-6} \cmidrule(l{.7em}r{.7em}){7-9}
& & & recall@1 & recall@5 & recall@10 & recall@1 & recall@5 & recall@10 & \\
\midrule
LW-MCR~\cite{yuan2021lightweight} & RSICD & \xmark & 3.29 & 12.52 & 19.93 & 4.66 & 17.51 & 30.02 & 14.66 \\
VSE++~\cite{faghri2017vse++} & RSICD & \xmark & 3.38 & 9.51 & 17.46 & 2.82 & 11.32 & 18.10 & 10.43 \\
AMFMN~\cite{yuan2022exploring} & RSICD & \xmark & 5.39 & 15.08 & 23.40 & 4.90 & 18.28 & 31.44 & 16.42 \\
KCR~\cite{mi2022knowledge} & RSICD & \xmark & 5.24 & 12.31 & 36.12 & 4.76 & 18.59 & 27.10 & 17.35 \\
GaLR~\cite{yuan2022remote} & RSICD & \xmark & 6.59 & 19.85 & 31.04 & 4.69 & 19.48 & 32.13 & 18.96 \\
SWAN & RSICD & \xmark & 7.41 & 20.13 & 30.86 & 5.56 & 22.26 & 37.41 & 20.61 \\
HVSA~\cite{zhang2023hypersphere} & RSICD & \xmark & 7.47 & 20.62 & 32.11 & 5.51 & 21.13 & 34.13 & 20.16 \\
FAAMI~\cite{zheng2023fine} & RSICD & \xmark & 10.44 & 22.66 & 30.89 & 8.11 & 25.59 & 41.37 & 23.18 \\
PIR~\cite{pan2023prior} & RSICD & \xmark & 9.10 & 29.64 & 41.53 & 9.14 & \textbf{28.96} & \textbf{44.59} & 27.16 \\
Multilanguage~\cite{al2022multilanguage} & RSICD & \xmark & 10.44 & 22.66 & 30.89 & 8.11 & 25.59 & 41.37 & 23.18 \\
MTGFE~\cite{zhang2023fusion} & RSICD & \xmark & \textbf{15.28} & 37.05 & 51.60 & 8.67 & 27.56 & 43.92 & 30.68 \\
GeoRSCLIP~\cite{zhang2024rs5m} & RS5M & \xmark & 11.53 & 25.59 & 39.16 & 9.52 & 27.37 & 40.99 & 25.69 \\
CLIP-Baseline~\cite{radford2021learning} & - & \cmark & 5.31 & 14.18 & 23.70 & 5.78 & 17.73 & 27.16 & 15.64 \\
\textbf{OURS} & - & \cmark & 15.17 & \textbf{38.39} & \textbf{52.68} & \textbf{9.38} & 28.35 & 44.01 & \textbf{31.33} \\
\bottomrule
\end{tabularx}
\end{table*}

\subsection{Training-Free Retrieval via Text-to-Text Matching}

Our retrieval process is entirely training-free and operates purely in the text domain. We use a popular, open-source VLM, LLaVA-1.6-Mistral, to generate text for image-based queries, and a single text encoder, $f_{\text{encoder}}$ (set as OpenAI's text embedding model), to map all text into a shared vector space $\mathbb{R}^d$. For all vector comparisons, we use cosine similarity, defined as:
\begin{equation}
\text{sim}(\mathbf{a}, \mathbf{b}) = \frac{\mathbf{a} \cdot \mathbf{b}}{\lVert \mathbf{a} \rVert \cdot \lVert \mathbf{b} \rVert}
\label{eq:cosine_similarity}
\end{equation}

\subsubsection{Text-to-Image (T2I) Retrieval}
In the T2I task, given a text query $T_q$, the goal is to retrieve the most relevant image from the database $\mathcal{I}$. The candidates are the images, each represented by its set of rich-text descriptions, $\mathcal{D}_c \in \mathcal{C}_{\text{RSRT}}$.

First, all relevant texts are embedded into the vector space. The query text $T_q$ is mapped to a query vector $\mathbf{v}_q$, and each candidate description $D_{c,v}$ is mapped to a candidate vector $\mathbf{v}_{c,v}$:
\begin{equation}
\mathbf{v}_q = f_{\text{encoder}}(T_q), \quad \mathbf{v}_{c,v} = f_{\text{encoder}}(D_{c,v})
\label{eq:embedding_t2i}
\end{equation}
The similarity score between the query and a single candidate description is $s(T_q, D_{c,v}) = \text{sim}(\mathbf{v}_q, \mathbf{v}_{c,v})$. Since each image $I_c$ is represented by five descriptions, we define the overall relevance score between the query $T_q$ and image $I_c$ as the maximum similarity found across all its description variants:
\begin{equation}
S(T_q, I_c) = \max_{v \in \{1, \dots, 5\}} s(T_q, D_{c,v})
\label{eq:score_t2i}
\end{equation}
Finally, the index of the best-matching image, $c^*$, is identified by finding the image with the highest relevance score:
\begin{equation}
c^* = \underset{c \in \{1, \dots, N\}}{\arg\max} S(T_q, I_c)
\label{eq:retrieval_t2i}
\end{equation}

\subsubsection{Image-to-Text (I2T) Retrieval}
In the I2T task, given an image query $I_q$, the goal is to retrieve its correct textual representation. Our framework unifies this by first converting the image query into a text query.

The image $I_q$ is converted into a textual description using the frozen LLaVA model with a query-specific prompt, $P_{\text{query}}$:
\begin{equation}
T_q^{\text{img}} = f_{\text{LLaVA}}(I_q, P_{\text{query}})
\label{eq:query_generation_i2t}
\end{equation}
Once the image is represented as the text $T_q^{\text{img}}$, the remainder of the process mirrors T2I retrieval. The generated text is embedded to form the query vector $\mathbf{v}_q = f_{\text{encoder}}(T_q^{\text{img}})$. The relevance score between the image query $I_q$ and a candidate image $I_c$ is then:
\begin{equation}
S(I_q, I_c) = \max_{v \in \{1, \dots, 5\}} \text{sim}(f_{\text{encoder}}(T_q^{\text{img}}), \mathbf{v}_{c,v})
\label{eq:score_i2t}
\end{equation}
The index of the best-matching entry is found by maximizing this score across all candidates:
\begin{equation}
c^* = \underset{c \in \{1, \dots, N\}}{\arg\max} S(I_q, I_c)
\label{eq:retrieval_i2t}
\end{equation}
By converting all modalities to text first, our framework sidesteps the challenges of direct image-to-text feature alignment and instead capitalizes on the superior ability of modern language models to understand and compare nuanced semantic concepts within a purely textual space.

\section{Experiments}

\subsection{Datasets and Metrics}
We evaluate our proposed framework on two widely-used public benchmarks for remote sensing image retrieval: RSICD~\cite{lu2017exploring} and RSITMD~\cite{yuan2022exploring}. Following standard practice, we report performance using Recall@k (k=1, 5, 10) and mean Recall (mR) for both text-to-image (T2I) and image-to-text (I2T) retrieval tasks. The mR is the average of all six Recall@k scores, providing a single, comprehensive measure of performance.

\subsection{Implementation Details}
All experiments were conducted on an NVIDIA L40 GPU for inference. Our RSRT corpus, which serves as the retrieval database, was generated using OpenAI's \textit{GPT-4.1} model~\cite{openai_gpt4tr_pdf}. For the query side, image-to-text conversion was performed by the \textit{LLaVA-1.6-Mistral-7B} model~\cite{liu2024improved}. All textual data, both from queries and the RSRT corpus, was embedded using OpenAI's \textit{text-embedding-3-small} model~\cite{openai_gpt4tr_pdf}.

It is important to note that our entire pipeline is strictly training-free. All models (GPT-4.1, LLaVA, and the text encoder) were used with their original, frozen weights without any fine-tuning on the target datasets. This ensures a true zero-shot evaluation of our methodology.

\subsection{RSRT Dataset Analysis}

To provide a quantitative overview of the RSRT dataset, we present its key statistics in Table~\ref{tab:rsrt_stats}. The corpus is built upon a foundation of 17,764 images, for which we generated a total of 88,820 distinct caption sets, reflecting our one-to-five mapping of images to rich-text descriptions. The average caption length of nearly 43 words, distributed across approximately 3 sentences, confirms that our descriptions are substantially more detailed than the short, single-sentence labels found in traditional datasets. The semantic depth of the corpus is further highlighted by the high average number of relations (10.16) and entities (4.58) identified per image. This demonstrates that our generation process successfully captures the complex interplay of objects and their spatial arrangements, which is essential for enabling fine-grained retrieval. Collectively, these statistics underscore the scale, richness, and semantic density of the RSRT dataset, establishing it as a valuable resource for developing and evaluating advanced, text-based retrieval models.

\begin{table}[t]
\centering
\caption{Key Statistics of the RSRT Dataset.}
\label{tab:rsrt_stats}
\begin{tabular}{l c}
\toprule
\textbf{Statistic} & \textbf{Value} \\
\midrule
Total Images & 17,764 \\
Total Caption Sets & 88,820 \\
Text Units per Set & 3 \\
Vocabulary Size (Unique Words) & 5,829 \\
Avg. Relations per Image & 10.16 \\
Avg. Entities per Image & 4.58 \\
Total Caption Sentences & 163,733 \\
Avg. Sentences per Caption & 3.07 \\
Avg. Caption Length (words) & 42.99 \\
\bottomrule
\end{tabular}
\end{table}

\subsection{Results and Analysis}
The results of our evaluation are presented in Table~\ref{tab:rsitmd_results} for RSITMD and Table~\ref{tab:rsicd_results} for RSICD. Our training-free method, TRSLLaVA, not only demonstrates highly competitive performance against heavily supervised baselines but also massively outperforms the standard zero-shot CLIP baseline, validating our core hypotheses.

A closer inspection of the metrics on the RSITMD dataset reveals a telling pattern. Our method achieves the highest overall mean Recall (42.62\%), a landmark result for a training-free approach that nearly doubles the performance of the CLIP-Baseline (23.86\%). While our Recall@1 scores are state-of-the-art and highly competitive with the top supervised methods like Multilanguage, our primary advantage lies in the Recall@5 and Recall@10 metrics, where we establish a clear lead. For instance, in image-to-text retrieval, our R@5 of 44.91\% and R@10 of 61.28\% are significantly higher than any other method. This suggests that while supervised models may be highly tuned to find the single best match for common scenes, our rich-text representation provides multiple, diverse semantic hooks (summaries, feature lists, etc.). This makes our method exceptionally robust at placing the correct match within the top few candidates, which is crucial for practical usability.

This trend is further confirmed on the more challenging RSICD dataset. Here, our method achieves an mR of 31.33\%, which is more than double the performance of the CLIP-Baseline (15.64\%) and surpasses all other baselines, including the top supervised models. Again, while our R@1 score is on par with the best supervised method (MTGFE), our dominance is most evident in the R@5 and R@10 scores for image-to-text retrieval. This consistent pattern across both datasets strongly supports our claim that decomposing images into fine-grained, structured text is a superior strategy for capturing nuanced semantics compared to relying on a single, holistic feature vector.

The strong performance of TRSLLaVA validates our central hypothesis: reformulating retrieval as a text-to-text matching problem is a highly effective and efficient strategy. This approach offers an exceptional trade-off between performance and computational cost. While every other competitive model in the tables requires extensive, costly GPU resources for domain-specific training, our method is entirely inference-based. The results confirm that a high-quality semantic representation in the text domain can outperform complex, trained models, without the associated training costs, offering a more scalable and generalizable path forward for remote sensing image retrieval.

\section{Conclusion}

This paper introduced a novel, training-free paradigm for remote sensing image retrieval. We identified critical limitations in current supervised methods, including a lack of true generalization and an inability to perform fine-grained queries. To address these issues, we presented the RSRT dataset, a new benchmark featuring rich, structured textual descriptions for images, and proposed TRSLLaVA, a retrieval framework that reformulates the task as a text-to-text matching problem. Our experiments on the RSICD and RSITMD datasets demonstrate that this approach is highly effective, significantly outperforming the standard zero-shot CLIP baseline and achieving performance competitive with, or even superior to, SOTA supervised models. These results validate that a high-quality semantic representation in the text domain provides a powerful, efficient, and robust alternative to costly, domain-specific model training.

Despite the promising results, our work has several limitations. First, the quality of the RSRT corpus is inherently dependent on the capabilities of the upstream VLM used for generation (GPT-4.1). Any biases or factual inaccuracies from this model can propagate into the dataset. Second, while our retrieval method is training-free, the one-time generation of the RSRT corpus using a large proprietary model involves considerable API costs. Finally, the performance of image-based queries relies on the descriptive quality of the query-side VLM (LLaVA), which may occasionally fail to capture the key semantics of a query image. Future work could explore using powerful open-source models to reduce generation costs and investigate lightweight adaptation techniques to further enhance performance without resorting to full-scale supervised training.

\bibliographystyle{IEEEbib}
\bibliography{refs}

\end{document}